\documentclass[11pt]{article}

\usepackage{amsmath,amsfonts,bm}

\def\eqref#1{equation~\ref{#1}}

\def\1{\bm{1}}

\DeclareMathAlphabet{\mathsfit}{\encodingdefault}{\sfdefault}{m}{sl}
\SetMathAlphabet{\mathsfit}{bold}{\encodingdefault}{\sfdefault}{bx}{n}

\usepackage{wrapfig,lipsum,booktabs}
\usepackage{hyperref}
\usepackage{url}
\usepackage{graphicx}
\usepackage{amsmath,amssymb,amsfonts}
\usepackage{pifont}
\usepackage{array}
\usepackage{multirow}
\usepackage[export]{adjustbox}
\usepackage{caption}
\usepackage{subcaption}
\usepackage{float}
\usepackage{newfloat}
\usepackage{makecell}
\usepackage{bbm}
\usepackage{bm}	
\usepackage{subcaption}
\usepackage{placeins}
\usepackage{listings}
\usepackage{bibentry}
\usepackage{arydshln}
\usepackage[noend]{algpseudocode}

\pdfoutput=1

\usepackage[]{ACL2023}
\usepackage{times}
\usepackage{latexsym}

\usepackage[T1]{fontenc}

\usepackage[utf8]{inputenc}

\usepackage{microtype}

\usepackage{inconsolata}

\title{PAD-Net: An Efficient Framework for Dynamic Networks}

\author{Shwai He\textsuperscript{\rm 1}\space\space
Liang Ding\textsuperscript{\rm 2}\thanks{~~Corresponding author}\space\space\space\space
Daize Dong\textsuperscript{\rm 3}\space\space 
Boan Liu\textsuperscript{\rm 4}\space\space 
Fuqiang Yu\textsuperscript{\rm 5}\space\space 
Dacheng Tao\textsuperscript{\rm 2}\\
    \textsuperscript{\rm 1}University of Maryland, College Park \space\space
    \textsuperscript{\rm 2}The University of Sydney\\
    \textsuperscript{\rm 3}University of Electronic Science and Technology of China\\
    \textsuperscript{\rm 4}Wuhan University\space\space 
    \textsuperscript{\rm 5}Shandong University\\
    {\tt\small shwaihe@umd.edu},\space\space
    {\tt\small liangding.liam@gmail.com}\space\space
}

\makeatletter
\newcommand{\biggg}{\bBigg@{1.2}}  
\def\bigggl{\mathopen\biggg}

\makeatother

\newcommand{\ignore}[1]{{}}

\begin{document}

\maketitle

\begin{abstract}
Dynamic networks, e.g., Dynamic Convolution (DY-Conv) and the Mixture of Experts (MoE), have been extensively explored as they can considerably improve the model's representation power with acceptable computational cost. The common practice in implementing dynamic networks is to convert the given static layers into fully dynamic ones where all parameters are dynamic (at least within a single layer) and vary with the input. However, such a fully dynamic setting may cause redundant parameters and high deployment costs, limiting the applicability of dynamic networks to a broader range of tasks and models. The main contributions of our work are challenging the basic commonsense in dynamic networks and proposing a partially dynamic network, namely PAD-Net, to transform the redundant dynamic parameters into static ones. Also, we further design Iterative Mode Partition to partition dynamic and static parameters efficiently. Our method is comprehensively supported by large-scale experiments with two typical advanced dynamic architectures, i.e., DY-Conv and MoE, on both image classification and GLUE benchmarks. Encouragingly, we surpass the fully dynamic networks by $+0.7\%$ top-1 acc with only $30\%$ dynamic parameters for ResNet-50 and $+1.9\%$ average score in language understanding with only $50\%$ dynamic parameters for BERT. Code will be released at: \url{https://github.com/Shwai-He/PAD-Net}. 
\end{abstract}

\section{Introduction}
\label{sec:introduction}

Deep neural networks have been continuously pushing the state-of-the-art performance in the tasks of computer vision~\citep{6909475,dosovitskiy2020image} and natural language processing~\citep{NIPS2015_7137debd,brunet2019understanding,zan-etal-2022-vega,zhong2022toward} in past years, at the cost of increasing training cost~\cite{shen2023efficient}. However, most prevalent architectures perform inference statically where both the computational graph and network parameters are fixed once after training, which limits the representation power. Dynamic networks \citep{han2021dynamic}, as opposed to static ones, adapt their parameters or architectures to each specific input, improving the model representation power with acceptable computational cost, e.g., Switch Transformers~\citep{fedus2021switch}. The common practice of implementing dynamic networks is transforming static networks (or modules) with counterpart dynamic ones. For example, Dynamic Convolution (DY-Conv) \citep{chen2020dynamic} replaces traditional convolution by adopting $k$ adaptive convolutional kernels; Mixture of Experts (MoE) \citep{DBLP:conf/iclr/ShazeerMMDLHD17} replaces a single fully connected layer with multiple feed-forward neural networks in a parallel manner.

The success of dynamic networks motivates practitioners to design dynamic networks \cite{fedus2021switch, li2021omni}, which often follow a fully dynamic approach, where all parameters are dynamic (at least within a single layer) and change with the input. Previous works \citep{chen2020dynamic} show that dynamic networks often outperform their static counterpart, and using more dynamic layers intriguingly leads to ever-increasing performance. For instance, dynamic convolution promotes the performance on the ImageNet when more static convolution layers turn into dynamic ones. However, these prior works do not explain the need for a fully dynamic mechanism, and it remains unclear whether to convert static parameters into dynamic and to what extent if yes. 

On the other hand, such a fully dynamic manner is resource expensive and may cause redundancy, limiting the applicability of dynamic networks. For instance, the total parameters of BERT-Base equipped with MoE are \textasciitilde 506.3M (with 8 experts) compared to only \textasciitilde 108.9M for vanilla BERT-Base. In addition, an MoE layer also multiplies the computation. It seems more is better when transforming static layers into dynamic ones, but how about the dynamic parameters within a dynamic network: Do all of them lead to the promotion? This urges us to reflect \textbf{\textit{whether there exist redundant dynamic parameters, in fully dynamic network layers? }} Based on the above scrutinization, we hypothesize that less is more for dynamic parameters in fully dynamic networks. 

Motivated by this hypothesis, we propose the Iterative Mode Partition (IMP) algorithm to progressively convert less important dynamic parameters into static ones for higher efficiency, while maintaining performance at a competitive level. Given a fully dynamic network initialized with all parameters in dynamic mode, we attempt to partition a subset of static parameters out from them. Specifically, we iteratively transform dynamic parameters based on their impact on loss values. If the transformation of the $i$-th element of dynamic parameters results in only a minimal loss difference, we safely make it static. Given a  desired dynamic ratio (the proportion of dynamic parameters), we can balance the trade-off between dynamic and static parameters. Since static parameters are less costly to deploy, we prune redundant parameters after mode partition, obtaining a lightweight architecture, namely Partially Dynamic Networks (\textbf{PAD-Net}), which contains two modes of parameters (dynamic parameters that vary with inputs and static parameters that are fixed during inference). 

Empirically, we extensively validate this hypothesis and our proposed PAD-Net, including GLUE benchmark~\citep{wang2018glue} for MoE and visual image classification~\citep{5206848} for dynamic convolution. Experiment results reveal that we successfully converted redundant dynamic parameters into static ones and PAD-Net achieves the highest performance in all tasks with lightweight architectures. Given the superiority of PAD-Net in both effectiveness and efficiency, we show that less dynamic is more efficient in fully dynamic networks, successfully verifying the above hypothesis. The inspiration of partially dynamic can be extended to other dynamic networks and even inform future efficient architectures designation. 

In short, our contributions are threefold: 

\begin{itemize}
\item[$\bullet$] We hypothesize that a fully dynamic network contains partial dynamic subnetworks that maintain or exceed the representation power of the original network.
\item[$\bullet$] Following our hypothesis, we propose the novel PAD-Net framework to achieve a partial dynamic mechanism and devise an \textit{Iterative Mode Partition} (IMP) algorithm to partition static and dynamic parameters. 
\item[$\bullet$] We empirically validate our hypothesis and PAD-Net on both NLP and CV tasks across two representative dynamic networks, including MoE and dynamic convolution. 
\end{itemize}
\section{Related Work}

\paragraph{Dynamic Networks.}
The dynamic neural network is an emerging research topic in deep learning, which adapts structures or parameters to different inputs, leading to notable advantages in terms of accuracy, and computational efficiency. \citet{han2021dynamic} classify dynamic networks into two categories: dynamic architecture networks and dynamic parameter networks. Dynamic architecture networks adaptively adjust architectures conditioned on each sample. Specifically, they adjust the network depth~\citep{wang2018skipnet}, width~\citep{mullapudi2018hydranets}, or route based on the input~\citep{huang2018condensenet}. Instead of changing the model architecture, dynamic parameter networks boost representation power by adapting parameters or activation functions to the input~\citep{yang2020condconv, liu2021dynamic}. Existing works often transform various types of static parameters into dynamic versions~\citep{chen2020dynamic}. Among them, dynamic convolution and mixture-of-experts are the typical examples that aggregate multiple convolution parameters (and experts) dynamically based on the input, leading to significant improvement with negligible computational cost. 

\paragraph{Network Pruning.} Past works in network pruning have explored effective techniques to find efficient subnetworks~\citep{lee2018snip, evci2020rigging, He2022SparseAdapterAE} and zero out redundant parameters. According to the lottery ticket hypothesis (LTH) pioneered by \citet{frankle2018lottery}, dense, randomly initialized, feed-forward networks contain the subnetwork (winning tickets) that maintains comparable test performance of the original network after training for the same iterations. This hypothesis inspires a series of follow-up works in network pruning. However, these methods always sacrifice performance because of pruned parameters. As for dynamic networks, instead of directly pruning dynamic parameters, we considered changing them to static ones. In Section~\ref{diff}, we show our approach significantly and consistently outperforms fully dynamic networks in the GLUE benchmark~\citep{wang2018glue}, while the pruned model performed worse than the original network.

\section{Review of Fully Dynamic Networks}
\label{:sec:Preliminaries}
\begin{figure*}[htbp]
  \centering
  \includegraphics[width=15.0cm]{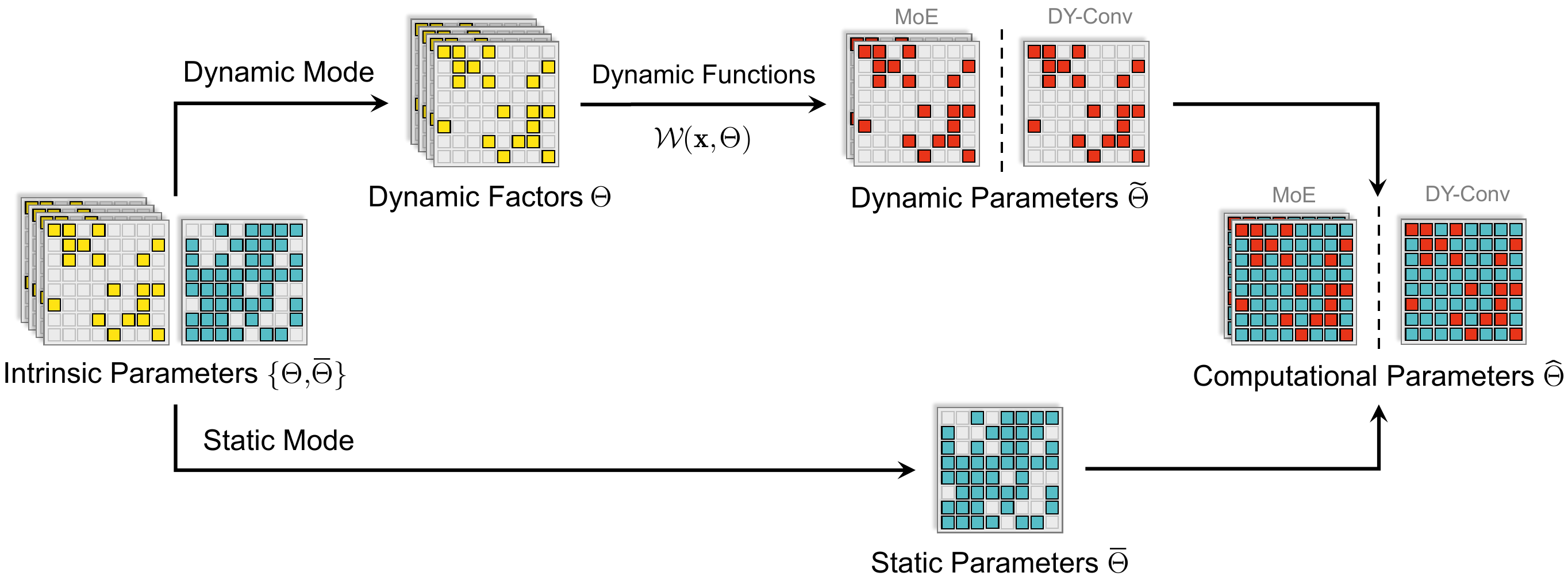}
  \vspace{8pt}
  \caption{\textbf{The procedure of generating the computational parameters in PAD-Net, with DY-Conv and MoE as instantiations.} The intrinsic parameters include static parameters and dynamic factors. Given an input, dynamic factors activate and aggregate into dynamic parameters, which are then integrated with static parameters.}
  \label{overview1}
  \centering
  \vspace{-10pt}
\end{figure*} 

\paragraph{Basic Concept.} Dynamic networks first adjust computational parameters and then compute the input using adjusted parameters, rather than directly using intrinsic parameters to compute the input. In a fully dynamic network, all intrinsic parameters are used as dynamic factors to generate computational parameters $\hat{{\Theta}}$, which are dependent on two parts: the input $\mathbf{x}$ and the intrinsic parameters $\Theta$. Let us denote ${\mathcal{W}}$ as the dynamic function, computational parameters is formulated as ${{\hat {\Theta}}} = {\mathcal{W}}(\mathbf{x}, {{\Theta}})$. Given an input sample $\mathbf{x}$, the output of is $\mathbf{y}=\mathcal{F}(\mathbf{x}, \Theta)$ for a conventional network with static parameters and $\mathbf{y}=\mathcal{F}(\mathbf{x}, \hat{{\Theta}})$ for a dynamic network. 

Existing dynamic networks, though using different dynamic functions, tend to follow a fully dynamic manner: Networks take all intrinsic parameters to generate the computational parameters where all elements are dynamic and vary with the input. We call such networks fully dynamic networks and, in the following, introduce instantiations coming from dynamic architecture networks, i.e., \textbf{\textit{Mixture of Experts}}, and dynamic parameter networks, i.e., \textbf{\textit{Dynamic Convolution}}, respectively.

\paragraph{Mixture of Experts.} We talk about dynamic architecture networks by taking the Mixture of Experts (MoE)~\citep{MoE1991, DBLP:conf/iclr/ShazeerMMDLHD17} as an instantiation. MoE prepares $m$ parallel static experts with parameters $\Theta^{(i)} (i=1,2,\dots,m) $ and only selects $n$ experts with the highest scores $(n \leq m)$. Given a specific input, we denote $G(\mathbf{x})$ as the output scores of gating and $\mathcal{T}$ as the indices of the selected experts. For the $i$-th selected expert, we denote the combination of the score $G_{\mathcal{T}_i}(\mathbf{x})$ and parameters $\Theta^{(\mathcal{T}_i)}$ as $w^{(\mathcal{T}_i)} = \left\{G_{\mathcal{T}_i}(\mathbf{x}), {\Theta}^{(\mathcal{T}_i)}\right\}$. The dynamic function of MoE can be represented as: 
\begin{equation}
\hspace{-10pt}
\label{moe}
\begin{aligned}
    \mathcal{W}(\mathbf{x}, {{\Theta}}) = \{w^{(\mathcal{T}_1)},\dots, w^{(\mathcal{T}_n)} \}, \\ 
    \text{where} \quad w^{(\mathcal{T}_i)} = \{G_{\mathcal{T}_i}(\mathbf{x}), {{\Theta}}^{(\mathcal{T}_i)}\}. 
\end{aligned}
\end{equation}

\paragraph{Dynamic Convolution.} As a typical example of dynamic parameter networks,  Dynamic Convolution~\citep{chen2020dynamic} prepares $k$ parallel static kernels ${{\Theta}}^{(i)} (i=1, 2, \dots,k)$ as intrinsic parameters and utilizes the linear combination of them as the aggregated kernel. The linear scale is aggregated dynamically via a channel-wise attention block~\citep{hu2018squeeze} denoted as $\operatorname{Attention}$, so the dynamic function can be written as: 
\begin{equation}
\hspace{-8pt}
\begin{aligned}
    {\mathcal{W}}(\mathbf{x}, {{\Theta}})=\sum_{i=1}^{k} {\pi}_{i}  (\mathbf{x}) \cdot {{\Theta}}^{(i)}, \\  
    \text{where} \quad {\pi}(\mathbf{x})= \operatorname{Attention} (\mathbf{x}).
\end{aligned}
\end{equation}
\paragraph{Limitation Discussions.} 
Mainstream dynamic networks usually replace static layers with fully dynamic layers, where all elements of dynamic parameters require corresponding dynamic factors co-working with input samples. 
However, this situation causes redundant parameters and high deployment costs, limiting the applicability of dynamic networks to a border range of resource-constrained situations and large-scale models. For this fully dynamic manner, we raise two questions: 
\textbf{(1) \textit{Is it necessary to pay the cost of enormous parameters and computations to aggregate dynamic parameters?}
(2) \textit{Is it necessary to make all computational parameters dynamic, to maintain the performance improvement?}} 
We propose the Partially Dynamic Network (PAD-Net) that mixes dynamic and static parameters to answer the above questions. 
\section{Methodology}
\label{sec:method}

\subsection{PAD-Net: Partially Dynamic Network}
\label{sec:pad}
In response to the limitation of fully dynamic networks, we question whether it is necessary to make all parameters dynamic. To this end, we try to detect the less important dynamic parameters and transform them into input-agnostic static parameters. Specifically, we utilize a mask ${\rm M}_i (i = 1,2, \dots, m)$ to indicate whether the $i$-th element of $\hat{{\Theta}}$ is dynamic or static: ${\rm M}_{i} = 1$ means the $i$-th element of $\hat \Theta$ is dynamic and vice versa. We use $\tilde{\Theta} \in \mathbb{R}^m$ to denote the dynamic parameters and  $\bar{\Theta} \in \mathbb{R}^m$ to represent the static parameters, then computational parameters ${\hat\Theta}$ are reformulated as:
\begin{equation}
\hspace{-14pt}
\label{eq1}
\hat{\Theta}_i = \begin{cases}
 \tilde {\Theta}_i = \mathcal{W}_i(\mathbf{x}, \Theta) & \text{if} \ \ {\rm M}_i = 1\\
 \bar{\Theta}_i & \text{otherwise}
\end{cases}, 
\end{equation}
where $\hat{{\Theta}}_i (i = 1,2,\dots,m)$ represents the $i$-th element of $\hat{{\Theta}}$, and $\Theta$ denotes the dynamic factors. In our architecture, intrinsic parameters include dynamic factors $\Theta$ and static parameters $\bar{\Theta}$. Note that ${\rm M}$ partitions the computational parameters into two non-overlapping parts, forming a network with only a part of the parameters dynamic, i.e., Partially Dynamic Network (PAD-Net). Details of the procedure of generating the computational parameters from intrinsic are visualized in Figure \ref{overview1}. 

To overcome the aforementioned challenges and limitations, we propose a novel network architecture, Partially Dynamic Network (PAD-Net). We also devise a new algorithm {\textit{Iterative Mode Partition} (IMP)} to build this model efficiently. 

In addition, we set two scale factors to describe the relative intensity of these subnetworks separately in terms of magnitude, namely $\mathbf{\lambda}_s$ and $\mathbf{\lambda}_d$. With these scale factors, we factorize our method into a more general formulation:
\begin{equation}
\label{eq10}
\hspace{-10pt}
\hat{{\Theta}}_i = \begin{cases}
 \mathbf{\lambda}_d \cdot \tilde{\Theta}_i & \text{if} \ \ {\rm M}_{i} = 1\\
 \mathbf{\lambda}_s \cdot \bar{\Theta}_i & \text{otherwise}
\end{cases},
\end{equation}
where we constrain $\mathbf{\lambda}_s + \mathbf{\lambda}_d = 2 (\mathbf{\lambda}_s, \mathbf{\lambda}_d > 0)$, and Equation \ref{eq1} is the special situation when both $\mathbf{\lambda}_s$ and $\mathbf{\lambda}_d$ are equal to 1. Similar to the constraint $\sum_{i=1}^{k} {\pi}_{i}$ in dynamic convolution \citep{chen2020dynamic}, this constraint compresses the parameters space and significantly simplifies the joint optimization of scale factors and the counterpart parameters. 

\subsection{Iterative Mode Partition}
\label{identify}

In the above section, we present the architecture of PAD-Net, which includes dynamic parameters and counterpart static parameters. Next, we further discuss our method in how to generate indicator masks to partition dynamic and static parameters. Let us first formulate this partition as an optimization problem, where our goal is to minimize loss values $L$. Given a dataset $ \mathcal{D} = \{(\mathbf{x}_i, \mathbf{y}_i)\}^n_{i=1}$ and a desired dynamic ratio $\kappa$ of ${\rm M}$, we briefly formulate mode partition as the following constrained optimization problem:
\begin{equation}
\small
\begin{aligned}
\min_{\rm M}
& L({\hat \Theta} , {\rm M} ; \mathcal{D}) = 
\min _{{\rm M}} \frac{1}{n}
\sum_{i=1}^{n} \ell ({\hat \Theta}, {\rm M}; 
(\mathbf{x}_{i}, \mathbf{y}_{i})), \\
& \text {s.t.} \quad {\rm M} \in \{0, 1\}^m, \hspace{2ex}  \|{\rm M}\|_{0} \leq \kappa \cdot m, 
\end{aligned}
\end{equation}
where $\ell(\cdot)$ denotes the standard loss function (e.g., cross-entropy loss), ${\hat \Theta}$ is the set of computational parameters of the neural network, $\Vert\cdot\Vert_{0}$ is the standard $L_0$ norm, $m$ is the total number of parameters. The conventional approach to optimize the above problem is adding sparsity enforcing penalty term ${\rm M}$~\citep{carreira2018learning}, while it often requires heavily tuned hyperparameter settings and several trials. On the other hand, LTH-based~\citep{chen2020lottery, evci2020rigging} methods find the mask by several iterations, but it is prohibitively time-consuming. Also, considering the large-scale dynamic networks, it is unnecessary to deploy redundant parameters. 

We tend to partition the two modes before training to prune redundant parameters and avoid time-consuming training iterations. Inspired by \citet{lee2018snip}'s gradient-based pruning strategy, we propose an algorithm to make excessive dynamic parameters static. We resort to mini-batches of training data $\mathcal{D}_b = {\{(\mathbf{x}_i, \mathbf{y}_i)\}}_{i=1}^b \sim \mathcal{D}$ to detect redundant dynamic parameters. Given a dynamic parameter ${\hat \Theta}_j$ at the $j$-th element of ${\hat \Theta}$, we compute its importance of being dynamic based on the loss difference $\Delta L_j$ caused by making ${\hat \Theta}_j$ static (changing the value of ${\rm M}_j$ from 1 to 0): 
\begin{figure*}[htbp]
  \centering
  \includegraphics[width=15.9cm]{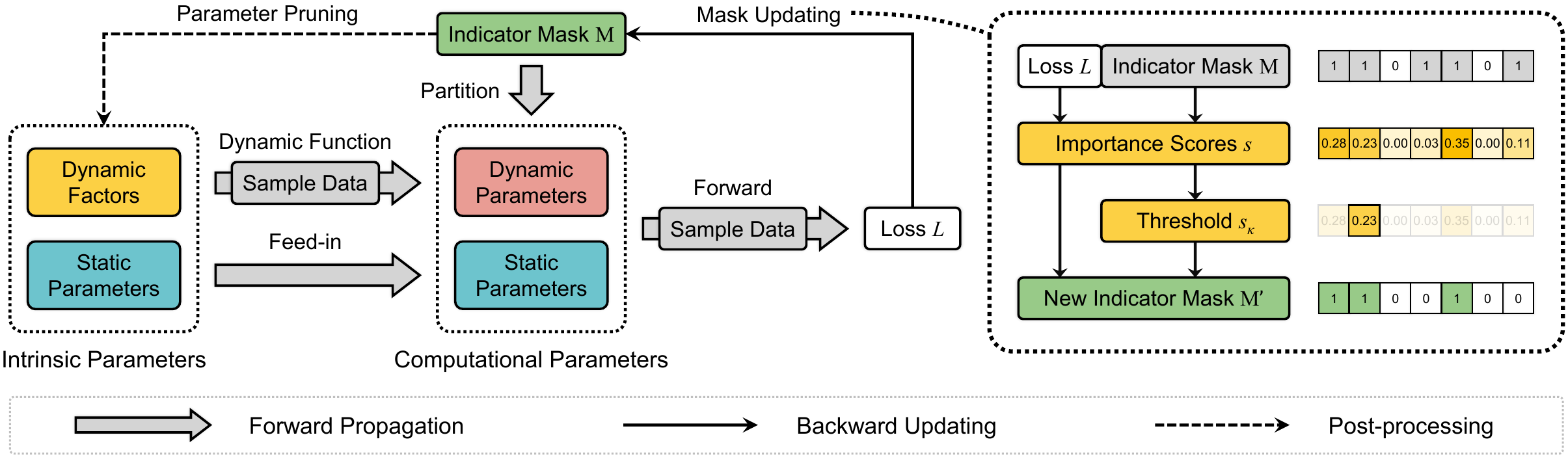}
  \vspace{-5pt}
  \caption{\textbf{Graphical illustration of Iterative Mode Partition (IMP).} \textbf{\em Left}: An overview of IMP, including forward propagation and backward updating. After IMP, the indicator mask prunes the redundant dynamic factors and static parameters (post-processing). \textbf{\em Right}: Details of mask updating.}
  \label{Mask_IMP}
  \vspace{-10pt}
\end{figure*} 
\begin{equation}
\small
\Delta L_j({\rm M}, {\hat \Theta}  ; \mathcal{D}_b) = L({\rm M}, {\hat \Theta}  ; \mathcal{D}_b)-L({\rm M} - \mathbf{t}_j, {\hat \Theta} ; \mathcal{D}_b),
\end{equation}
where $\mathbf{t}_j$ is the indicator vector of $j$-th element of ${\rm M}$  (i.e., zeros everywhere except at the index $j$ where it is one). We only consider transforming redundant dynamic parameters into static ones, so the loss difference $\Delta L_j$ is zero if ${\hat \Theta}_j$ is static. 
Note that computing $\Delta L_{j}$ for each dynamic parameter is prohibitively expensive, as it usually requires millions of forwarding passes over the dataset, so we resort to a simple and effective approximate alternative. Specifically, we release binary constraints of ${\rm M}$ and make it differentiable and utilize the derivative of $L$ with respect to ${\rm M}_j$ to approximate $\Delta L_j$: 
\begin{equation}
\small
\begin{aligned}
\Delta & L_j ({\rm M}, {\hat \Theta}  ; \mathcal{D}_b) 
 \approx g_j({{\hat \Theta}} ; \mathcal{D}_b)
= \frac{\partial L_j({\rm M}, \hat \Theta  ; \mathcal{D}_b)}{\partial {\rm M}} \\
&= \left.\lim _{\delta \rightarrow 0} \frac{L_j({\rm M}, \hat \Theta  ; \mathcal{D}_b) - L_j({\rm M} - \delta \mathbf{t}_j, \hat \Theta  ; \mathcal{D}_b)}{\delta}\right|_{\mathbf{t=1}}, 
\end{aligned}
\end{equation}
where $g_j(\hat \Theta ; \mathcal{D}_b)$ denotes the $j$-th element in derivative $g(\hat \Theta ; \mathcal{D}_b)$. We accumulate the derivatives for all $j$ by one forward-backward pass using automatic differentiation. Note that if the magnitude of $g_j$ is high, it essentially means that making parameter $\hat \Theta_j$ static has a considerable effect on the loss, and it has to be dynamic. In contrast, the parameter should be static if the magnitude of $g_j$ is low. Therefore, We take the normalized magnitude of the derivatives of $g$ as the criteria:
\begin{equation}
        s_j = {\left|g_j(\hat \Theta ; \mathcal{D}_b)\right|}\bigggl /{\sum\nolimits_{k=1}^{m}\left|g_k (\hat \Theta ; \mathcal{D}_b)\right|}. 
\end{equation}
Given the dynamic ratio $\kappa$, we take the $s_{\kappa}$ (the $\kappa$-th percentile of $s$) as the threshold and transform the mask elements whose scores are below zero:
\begin{equation}
    {\rm M} = \mathbbm{1} \left[s - s_{\kappa} \geq 0\right], 
\end{equation}
where $\mathbbm{1}[\cdot]$ is an element-wise indicator function where the output will be 1 if the condition $[\cdot]$ meets else it will be zero. Note that the indicator mask ${\rm M}$ prunes out redundant parameters in dynamic parameters $\tilde{{\Theta}}$ and static parameters $\bar{{\Theta}}$ respectively. Also, for fewer dynamic parameters to generate, we can also utilize the binary mask to prune redundant dynamic factors.  
Taking MoE as an example, $\rm M$ can be directly applied to parallel experts: $\Theta^{(i)} \leftarrow {\rm M} \odot \Theta^{(i)}, \forall i \in \{1,2, \dots, k\}$. In addition, we can decrease the computational cost of generating based on dynamic factors. 

Inspired by the success of iterative strategy in pruning at initialization~\citep{verdenius2020pruning, de2020progressive}, we start from a fully dynamic network and adopt an iterative strategy shown in Figure \ref{Mask_IMP} to transform dynamic parameters into static parameters step by step, where we increase the zero ratio of $\rm M$ exponentially. The effectiveness of the mode partition and the iterative mode partition are experimentally verified in Section~\ref{eff_imp}. 
\section{Epirical Evaluation}
\label{sec:empirical}

\begin{table*}[t]
\centering
\caption{\textbf{Comparison between PAD-Net and vanilla MoE} applied to four widely used large-scale Pretrained Language Models (PLMs). Averaged scores on all tasks are \underline{underlined}. The shown results are the averaged score for 5 runs, followed by the deviation. The best results are \textbf{bold}. It shows that PAD-Net yields consistent improvements across all tasks among different MoE-equipped PLMs. 
}
\resizebox{1.00\textwidth}{!}{
\begin{tabular}{ccccccccccccc}
    \toprule
    \multirow{2}{*}{\bf Method} & \multicolumn{5}{c}{BERT} & \multicolumn{7}{c}{ALBERT} \\ \cmidrule(lr){2-7} \cmidrule(lr){8-13}
    ~ & \#Param. & CoLA & RTE & MRPC & STS-B & \underline{Avg.} & \#Param. & CoLA & RTE & MRPC & STS-B & \underline{Avg.}  \\
    \midrule
    Static & 108.9M & $54.6_{\pm 0.4}$ & $66.4_{\pm 0.7}$ & $84.6_{\pm 0.3}$ & $85.8_{\pm 0.3}$ & \underline{72.9} & 
    11.1M & $54.2_{\pm 0.7}$ & $76.6_{\pm 0.7}$ & $87.2_{\pm 0.4}$ & $90.6_{\pm 0.3}$ & \underline{77.2} \\
    \hdashline
    MoE & 506.3M & $58.0_{\pm 0.9}$ & $69.3_{\pm 1.2}$ & $85.0_{\pm 0.4}$ & $87.1_{\pm 0.2}$ & \underline{74.9} & 
    44.2M & $56.8_{\pm 1.2}$ & $77.2_{\pm 0.8}$ & $87.4_{\pm 0.4}$ & $90.7_{\pm 0.3}$ & \underline{78.0} \\
    
    PAD-Net & \textbf{308.0M} & $\textbf{59.7}_{\pm 0.8}$ & $\textbf{71.5}_{\pm 1.4}$ & $\textbf{85.5}_{\pm 0.4}$ & $\textbf{90.3}_{\pm 0.6}$ & \textbf{\underline{76.8}} & 
    \textbf{30.0M} & $\textbf{57.4}_{\pm 1.4}$ & $\textbf{77.6}_{\pm 0.5}$ & $\textbf{88.4}_{\pm 0.3}$ & $\textbf{90.9}_{\pm 0.2}$ & \textbf{\underline{78.6}} \\
    \midrule
    \multirow{2}{*}{\bf Method} & \multicolumn{5}{c}{RoBERTa} & \multicolumn{7}{c}{ELECTRA} \\ \cmidrule(lr){2-7} \cmidrule(lr){8-13}
    ~ & \#Param. & CoLA & RTE & MRPC & STS-B & \underline{Avg.} & \#Param. & CoLA & RTE & MRPC & STS-B & \underline{Avg.}  \\
    \midrule
    Static  & 124.1M & $62.8_{\pm 1.0}$ & $77.6_{\pm 1.6}$ & $90.0_{\pm 0.5}$ & $91.0_{\pm 0.3}$ & \underline{80.4} &
    108.9M & $67.3_{\pm 1.5}$ & $82.6_{\pm 1.7}$ & $89.0_{\pm 0.5}$ & $90.6_{\pm 0.1}$ & \underline{82.4} \\
    \hdashline
    
    MoE  & 521.5M & $63.6_{\pm 1.1}$ & $78.0_{\pm 1.4}$ & $90.2_{\pm 0.4}$ & $91.1_{\pm 0.2}$ & \underline{80.8} & 
    506.3M & $67.6_{\pm 1.1}$ & $83.0_{\pm 1.4}$ & $89.3_{\pm 0.3}$ & $90.8_{\pm 0.2}$ & \underline{82.7} \\
    
    PAD-Net  & \textbf{323.1M} & $\textbf{64.2}_{\pm 0.8}$ & $\textbf{79.4}_{\pm 1.2}$ & $\textbf{90.7}_{\pm 0.3}$ & $\textbf{91.4}_{\pm 0.3}$ & \textbf{\underline{81.4}} & 
    \textbf{308.0M} & $\textbf{68.2}_{\pm 1.3}$ & $\textbf{84.1}_{\pm 1.5}$ & $\textbf{89.5}_{\pm 0.4}$ & $\textbf{91.2}_{\pm 0.2}$ & \underline{\textbf{83.3}} \\
    \bottomrule
    \end{tabular}
}
\label{tab:glue}
\end{table*}

\begin{table*}[htbp]
\centering
\caption{\textbf{Comparison between PAD-Net and baselines for ResNet and MobileNetV2}, including CondConv and DY-Conv. The Top-1 accuracy is the averaged score for 5 runs, followed by the deviation. \ding{86} indicates the dynamic model with the fewest parameters or the fewer FLOPs (the static model is not included), and the best results in accuracy are \textbf{bold}. DY-Conv and PAD-Net contain $k=4$ kernels, while CondConv contains $k=8$ kernels.}

\begin{minipage}{0.48\linewidth}
\centering
\begin{adjustbox}{max width=1.00\linewidth}
  \begin{tabular}{clrrc}
  \toprule
  Depth & Model & Params & FLOPs & \makecell[l]{Top-1(w/dev)} \\
  \midrule
  \multirow{4}{*}{ResNet-10} 
  & Static & 5.2M & 0.89G    & $63.1_{\pm 0.4}$ \\
  & CondConv & 36.7M & 0.92G & $66.9_{\pm 0.2}$\\
  & DY-Conv & 18.6M & 0.91G  & $67.4_{\pm 0.3}$ \\
  & PAD-Net & \textbf{\ding{86}6.9M} & \textbf{\ding{86}0.90G}  & $\textbf{68.1}_{\pm 0.2}$ \\
  \midrule
  \multirow{4}{*}{ResNet-18} 
  & Static & 11.1M & 1.81G   & $70.6_{\pm 0.3}$ \\
  & CondConv & 81.4M & 1.89G & $71.9_{\pm 0.2}$ \\
  & DY-Conv & 42.7M & 1.86G  & $72.4_{\pm 0.3}$ \\
  & PAD-Net & \textbf{\ding{86}15.1M} & \textbf{\ding{86}1.83G} & $\textbf{73.0}_{\pm 0.3}$ \\
  \midrule
  \multirow{4}{*}{ResNet-50} 
  & Static & 23.5M & 3.86G & $76.2_{\pm 0.2}$ \\
  & CondConv & 129.9M &3.98G & $76.9_{\pm 0.3}$ \\
  & DY-Conv & 100.9M & 3.97G& $77.2_{\pm 0.2}$ \\
  & PAD-Net & \textbf{\ding{86}33.8M} & \textbf{\ding{86}3.90G}  & $\textbf{77.9}_{\pm 0.2}$ \\
  \toprule
  \end{tabular}
  \end{adjustbox}
        \subcaption{\normalsize  ResNet}
\end{minipage}
\hspace{.10in}
\begin{minipage}{0.46\linewidth} 
\centering
\begin{adjustbox}{max width=1.00\linewidth}
  \begin{tabular}{clrrc}
  \toprule
  Width & Model & Params & FLOPs & \makecell[l]{Top-1(w/dev)} \\
  \midrule
  \multirow{4}{*}{\makecell[c]{$\times 0.5$}} 
  & Static & 2.0M    & 97.0M         & $65.7_{\pm 0.3}$ \\
  & CondConv & 15.5M & 113.0M        & $68.8_{\pm 0.2}$ \\
  & DY-Conv & 4.0M   & 101.4M        & $69.6_{\pm 0.1}$ \\
  & PAD-Net & \textbf{\ding{86}2.7M} & \textbf{\ding{86}98.3M} & $\textbf{70.4}_{\pm 0.2}$ \\
  \midrule
  \multirow{4}{*}{\makecell[c]{$\times 0.75$}} 
  & Static & 2.6M    & 209.1M       & $69.2_{\pm 0.4}$ \\
  & CondConv & 17.5M & 233.9M       & $72.1_{\pm 0.3}$ \\
  & DY-Conv & 8.0M   & 220.1M       & $72.6_{\pm 0.1}$ \\
  & PAD-Net & \textbf{\ding{86}5.2M} & \textbf{\ding{86}212.4M} & $\textbf{73.5}_{\pm 0.2}$ \\
  \midrule
  \multirow{4}{*}{\makecell[c]{$\times 1.0$}} 
  & Static & 3.5M    & 300.8M        & $72.1_{\pm 0.3}$ \\
  & CondConv & 27.5M & 329.0M        & $74.4_{\pm 0.2}$\\
  & DY-Conv & 11.1M  & 312.9M        & $74.8_{\pm 0.2}$ \\
  & PAD-Net & \textbf{\ding{86}6.1M} & \textbf{\ding{86}304.4M} & $\textbf{75.3}_{\pm 0.1}$ \\
  \toprule
  \end{tabular}
  \end{adjustbox}
        \subcaption{\normalsize  MobileNetV2}
\end{minipage}
\label{dyconv}
\vspace{-10pt}
\end{table*}

\subsection{Implementation Details}
\label{imple}
\paragraph{Mixture of Experts.} We use Adam~\citep{kingma2014adam} as the optimizer with $\beta_1$, $\beta_2$ = 0.9, 0.98. For regularization, we set the weight decay as 0.1 and grid-search the learning rate from \{1e-5, 5e-5, 1e-4, 5e-4\}, where we warm up the learning rate in the first 10\% steps (of the total training steps). For different data scales, we grid-search training epoch and batch size from \{5, 10, 15, 20\} and \{8, 16, 32, 64\}, respectively. The maximum length is 128 for all tasks. Following \citet{DBLP:conf/iclr/ShazeerMMDLHD17}, we initialize dynamic and static parameters with pretrained parameters. 

\paragraph{Dynamic Convolution.} We use an SGD optimizer \citep{ruder2016overview} with 0.9 momentum, following cosine learning rate scheduling and warmup strategy.  The learning rate rises to the maximum linearly in the first ten epochs and schedules to arrive at zero within a single cosine cycle. We follow~\citet{chen2020dynamic}'s temperature annealing strategy to avoid the unstable output values of the softmax function in the first epochs. We train ResNet for 100 epochs with the max learning rate of 0.1. We train the MobilenetV2 for 300 epochs with the max learning rate of 0.05. The weight decay is 1e-4 for ResNet and 4e-5 for MobilenetV2. The training batch size is 256 for all models. 

\subsection{Main Results}
\paragraph{Natural Language Understanding. }
We evaluate the performance of PAD-Net for MoE on various datasets from the General Language Understanding Evaluation (GLUE) benchmark~\citep{wang2018glue}.
Like previous works~\citep{lee2019mixout, dodge2020fine,zhong2022improving}, we fine-tune pretrained models, e.g., BERT \citep{devlin2018bert}, ALBERT \citep{lan2019albert}, RoBERTa \citep{liu2019roberta}, ELECTRA \citep{clark2020electra} on the training set and directly report results on validation set using the last checkpoint, since the test results are only accessible by the leaderboard with submission limitation.

\begin{table*}
\centering
\caption{\textbf{Ablation study for dynamic ratio on MoE integrated with PAD-Net.} Averaged scores on all tasks are \underline{underlined}. The shown results are the averaged score for 5 runs. The best results are \textbf{bold}. Methods under the dashline are our proposed PAD-Net, where $\kappa$ denotes the dynamic ratio. }
\resizebox{1.00\linewidth}{!}{
\begin{tabular}{lccccccccccc}
    \toprule
    \bf Method~~ & ~~\#Param.~~ & ~~CoLA~~ & ~~SST-2~~ & ~~MRPC~~ & ~~STS-B~~ & ~~QQP~~ & ~~MNLI~~ & ~~QNLI~~ & ~~RTE~~ & ~~\underline{Avg.}~~ \\
    \midrule
    BERT          & 108.9M & 54.6 & 91.4 & 84.6 & 85.8 & 90.6 & 83.7 & 90.4 & 66.4 & \underline{81.2} \\
    w/ MoE        & 506.3M & 58.0 & 91.7 & 85.0 & 87.1 & 90.8 & 83.8 & 90.8 & 69.3 & \underline{82.1} \\
    \hdashline
    $\kappa = 70\%$ & 387.3M & 58.5 & \bf 92.4 & \bf 85.5 & 89.6 & 90.9 & 83.9 & 90.9 & 70.6 & \underline{82.8} \\
    $\kappa = 50\%$ & 308.0M & \bf 59.7 & 92.2 & 85.4 & \bf 90.3 & 90.9 & \bf 84.2 & \bf 91.0 & \bf 71.5 & \bf \underline{83.2} \\
    $\kappa = 30\%$ & 228.6M & 59.0 & 92.0 & 85.3 & 89.4 & \bf 91.0 & 84.0 & 90.9 & 71.2 & \underline{82.9} \\
    $\kappa = 10\%$ & 149.3M & 57.5 & 91.1 & 85.4 & 88.3 & 90.4 & 83.6 & 90.6 & 70.2 & \underline{82.1} \\
    \midrule
    RoBERTa       & 124.1M & 62.1 & 94.0 & 89.6 & 90.6 & 91.0 & 86.9 & 91.8 & 77.4 & \underline{85.4} \\
    w/ MoE        & 521.5M & 63.6 & 94.8 & 90.2 & 91.1 & 91.7 & 87.7 & 92.9 & 78.0 & \underline{86.3} \\
    \hdashline
    $\kappa = 70\%$ & 402.5M & \bf 64.6 & 95.0 & \bf 91.0 & 91.0 & 91.8 & 87.7 & \bf 92.9 & 78.2 & \underline{86.5} \\
    $\kappa = 50\%$ & 323.1M & 64.4 & \bf 95.2 & 90.7 & \bf 91.4 & \bf 91.9 & \bf 88.0 & 93.0 & \bf 79.4 & \bf \underline{86.8} \\
    $\kappa = 30\%$ & 243.8M & 63.4 & 94.6 & 90.5 & 91.2 & 91.4 & 87.8 & 93.2 & 78.8 & \underline{86.4} \\
    $\kappa = 10\%$ & 164.5M & 63.9 & 94.4 & 90.4 & 90.8 & 90.9 & 87.4 & 92.6 & 78.2 & \underline{86.1} \\
    \bottomrule
    \end{tabular}
}
\label{tab:dynamic ratio moe}
\end{table*}

\begin{figure}[ht]
    \caption{\textbf{Normalized performance of ResNet with different dynamic ratios}, which is evaluated by $\frac{x - \bar{x}}{\bar{x}}$ where $\bar{x}$ is the mean accuracy across experiments. }
\includegraphics[width=0.96\linewidth]{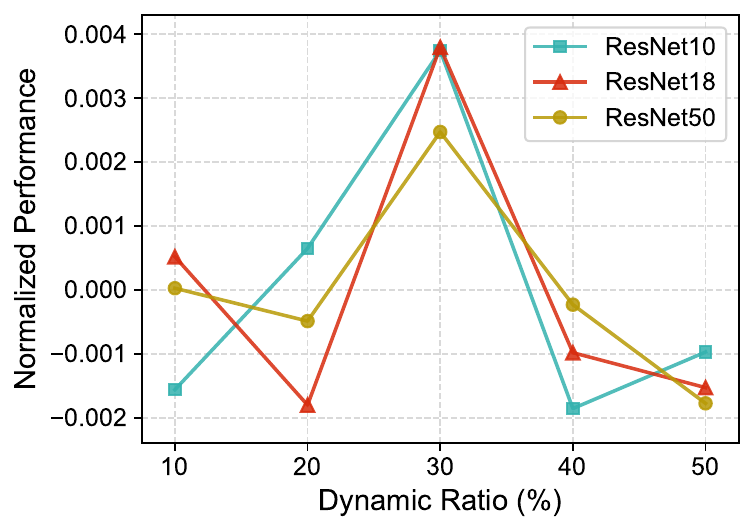}
  \label{fig: dynamic ratio}
\end{figure}

Following \citet{DBLP:conf/iclr/ShazeerMMDLHD17, gao-etal-2022-parameter}, we replace feed-forward layers with MoE layers where we prepare 8 experts and select the top-2 experts for each input. We set the dynamic ratio $\kappa=50\%$ because it is close to the optimal value.  Table \ref{tab:glue} shows that PAD-Net outperforms MoE on the GLUE benchmark with a 0.95\% average increase for four backbones. Specifically, PAD-Net improves BERT by 1.9\% and RoBERTa by 0.6\% on average. Equipped with PAD-Net, MoE reduces both parameters and computation significantly, and we provide a theoretical analysis of the reduced computation in Section~\ref{analysis: computation}. 

\begin{table}[ht]
\centering
  \caption{\textbf{Ablation study of scale factors, } where ``Option'' refers to the setting of scale factors. }
  \begin{adjustbox}{max width=0.94\linewidth}
  \begin{tabular}{ccccc}
    \toprule
     Model & Option & RTE & STS-B \\
    \midrule
  \multirow{5}{*}{\makecell{\text{BERT}-{\text{base}}}} & -- & 69.6 & 87.4 \\  
          & $\lambda_s$ & 70.7 & 88.1 \\ 
          & $\lambda_d$ & 70.9 & 89.6 \\
          & $\lambda_s, \lambda_d$ & 71.3 & 89.8 \\
          & $\lambda_s + \lambda_d = 2$ & \bf 71.5 & \bf 90.3 \\ 
\midrule
  Model & Option & CIFAR-10  & ImageNet \\ 
  \midrule
  \multirow{5}{*}{ResNet-50}& -- & 93.9 & 77.1 \\ 
          & $\lambda_s$ & 94.3 & 77.2 \\
          & $\lambda_d$ & 94.5 & 77.4 \\ 
          & $\lambda_s, \lambda_d$ & 96.0 & 77.6 \\
          & $\lambda_s + \lambda_d = 2$ & \bf 96.6 & \bf 77.8 \\ 
    \toprule
  \end{tabular}
  \end{adjustbox}
  \label{scale}
\end{table}

\paragraph{Visual Image Classification.}
We also report the superiority of PAD-Net in visual image classification. In Table \ref{dyconv}, we compare PAD-Net with static convolution \citep{krizhevsky2012imagenet}, CondConv \citep{yang2020condconv} and Dynamic Convolution \citep{chen2020dynamic} on ImageNet \citep{5206848} classification for ResNet \citep{he2016deep} and MobileNetV2 \citep{sandler2018mobilenetv2} in the same experimental setting with previous works, by adjusting all convolution layers except the first layer. Before training, we first partition two modes of parameters with a given dynamic ratio $\kappa$ using ten batches of examples. 

PAD-Net improves accuracy with significantly lighter architecture and fewer FLOPs (Floating Point Operations). For instance, PAD-Net outperforms DY-Conv by 0.7\% top-1 accuracy with 33.9\% parameters and 0.1G fewer FLOPs in ResNet-50. 


\subsection{Ablation Study}

\paragraph{Effect of Dynamic Ratio. } 
Inspired by~\citet{wettig2022should}, we investigate the impact of different dynamic ratios $\kappa$, and the results are shown in Table \ref{tab:dynamic ratio moe} For MoE and Figure \ref{fig: dynamic ratio} for DY-Conv. Because PAD-Net with low dynamic ratios significantly outperforms fully dynamic networks, we only consider ratios of less than 70\%, allowing for more sparsity and efficiency. As shown, we empirically find that $\kappa=50\%$ is nearly the optimal ratio for MoE to achieve the highest performance, where the best performance of DY-Conv is achieved when $\kappa = 30\%$. We believe that different dynamic functions contribute to different optimal dynamic ratios, and an efficient way, e.g., hyper-parameter optimization or meta-learning, to search them will be necessarily explored in the future. 

\begin{figure}
  \centering
  \caption{\textbf{Comparison of different partition methods}, including random partition ``Random'', mode partition ``MP'', and iterative mode partition ``IMP''. We also report dynamic convolution ``Dynmiac'' as a baseline. }
  \hspace{-5pt}
  \includegraphics[width=0.475\textwidth]{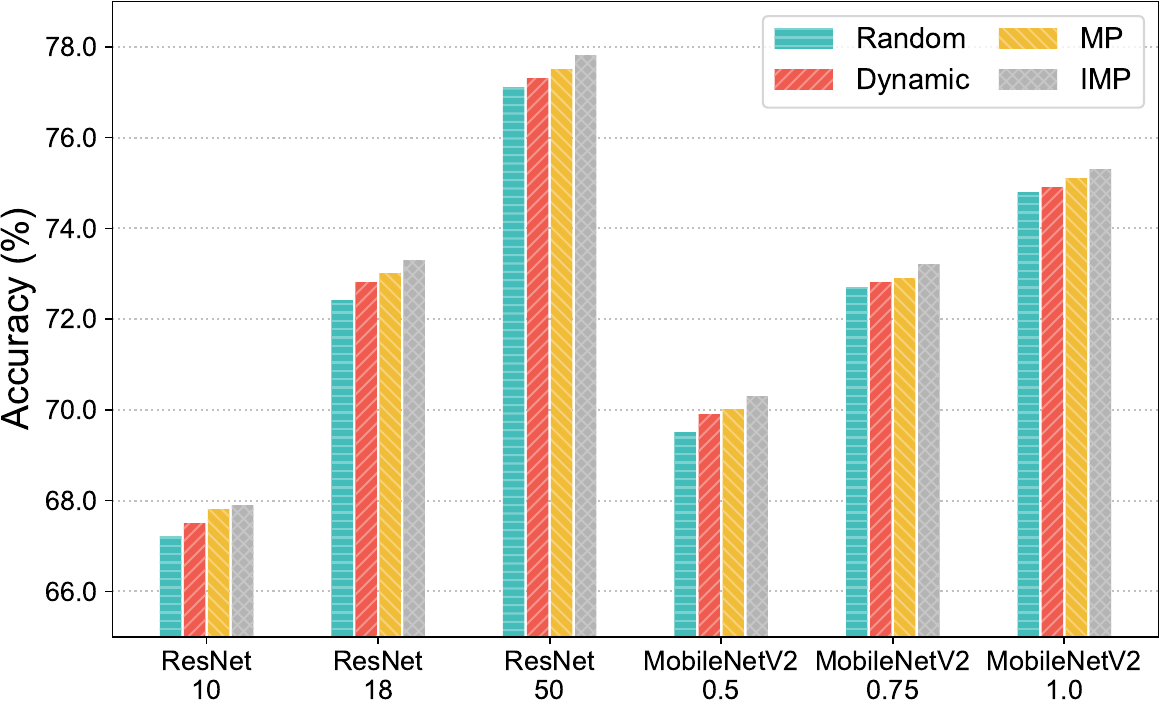}
  \label{imp-effect}
\end{figure}

\paragraph{Effect of Scale Factors.} 
We also conduct an ablation study on the proposed scale factors and verify their necessity. Table \ref{scale} summarizes the impact of scale factors on different architectures. We initially tried to gain scale factors from a SENet structure \citep{hu2018squeeze}, while it did not contribute to the improvement of performance. So we just set scale factors as trainable parameters to avoid redundant parameters and operations. Besides the setting ``$\lambda_s$+$\lambda_d = 2$'' in Equation~\ref{eq10}, we consider other situations: only using one factor (``$\lambda_s$'' and ``$\lambda_d$'') , and no scale factors used (``--''). We conduct experiments on CIFAR-10 \citep{Krizhevsky09learningmultiple} and ImageNet for ResNet-50, RTE, and STS-B for BERT. $\lambda_s$ and $\lambda_d$ enhance performance substantially, and their coexistence leads to further improvement. To explore the impact of the summation constraint, we release it and denote this setting as ``$\lambda_s, \lambda_d$''. Clearly, without summation constraint, the performance of ResNet-50 and BERT decreases significantly, i.e., -0.4\% and -0.35\% on average. 

\paragraph{Effectiveness of Iterative Mode Partition.}
\label{eff_imp}
We compare different partition strategies in Figure \ref{imp-effect}. Compared to fully dynamic networks, accuracy degrades when we partition two modes randomly, which means this naive partition method mistakes some important dynamic parameters. In contrast, mode partition contributes to a better combination of dynamic and static parameters, improving the accuracy. IMP shows its effectiveness by achieving the best performance. 

\subsection{Detailed Analysis}
\paragraph{Reduced Computation.}
\label{analysis: computation}

\begin{figure}
  \centering
  \caption{\textbf{Visualization description of the computation cost for PAD-Net on MoE. } Given a specific input $X$, we denote the computation cost for selected experts and static parameters. }
  \includegraphics[width=0.47\textwidth]{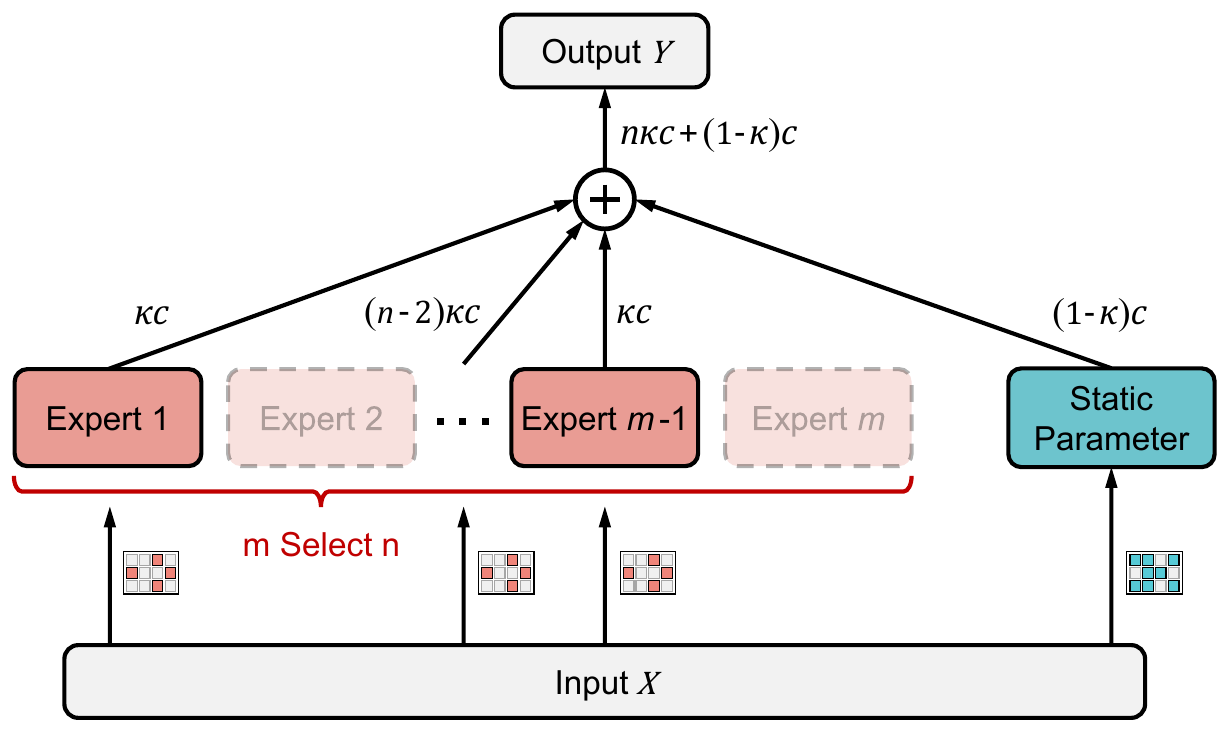}
  \label{computation}
\end{figure}

We show the computation cost of PAD-Net in Figure \ref{computation}. Compared to vanilla MoE, PAD-Net reduces the computation between selected experts and the input, $y_{\mathcal{T}_i} = E_{\mathcal{T}_i} (x)$, where $E_{\mathcal{T}_i}$ denotes the $i$-th selected experts. Because the two methods share the same gating mechanism, we temporally ignore its computation for simplicity. We denote the computation of the $i$-th expert as $C_{\mathcal{T}_i}$ where $C_{\mathcal{T}_1} = \dots = C_{\mathcal{T}_n} = c$, and the total computation of multi-experts is $n c$ if we select $n$ experts within $m$ ones. In PAD-Net, given the dynamic ratio $\kappa$, it is reduced to $n \kappa c$. Together with the computation $(1 - \kappa) c$, the computation of a PAD-Net layer is $n \kappa c + (1 -  \kappa)c$. Integrated with PAD-Net, an MoE layer can reduce the computation by $(n - 1)(1 - \kappa)c$. When $\kappa$ is low enough, the computation of PAD-Net can be close to static networks. For DY-Conv, the reduced computation lies in the linear combination of parallel kernels $\sum_{i=1}^{k} {\pi}_{i}  (\mathbf{x}) \cdot {{\Theta}}^{(i)}$, which is sparse in PAD-Net. In short, the degree of reduced computation depends on the specific dynamic function.

\begin{table*}
\centering
\caption{\textbf{Empirical comparison between our PAD-Net and model pruning on the GLUE benchmark}. PAD-Net is set with $\kappa = 50\%$, and MoE-P is an MoE architecture pruned by SNIP \cite{lee2018snip}. We make the parameters of PAD-Net and MoE-P consistent for a fair comparison. 
}
\resizebox{\linewidth}{!}{ 
\begin{tabular}{lcccccccccccc}
\toprule
\bf Method~~ & ~~\#Param.~~ & ~~CoLA~~ & ~~SST-2~~ & ~~MRPC~~ & ~~STS-B~~ & ~~QQP~~  & ~~MNLI~~  & ~~QNLI~~ & ~~RTE~~  & ~~\underline{Avg.}~~  \\
\midrule
Static & 108.9M & $54.6$ & $91.4$  & $84.6$ & $85.8$ & $90.6$ & $83.7$ & $90.4$ & $66.4$ & 81.2 \\
MoE & 506.3M & $58.0$ & $91.7$  & $85.0$ & $87.1$ & $90.8$ & $83.8$ & $90.4$ & $69.3$ & 82.0 \\
\hdashline
MoE-P & \textbf{\multirow{2}{*}{308.0M}} & $55.6$ & $91.6$ & $84.7$ & $85.8$ & $90.8$ & $82.4$ & $90.2$ & $65.7$ & 80.9 \\
PAD-Net & & $\textbf{59.7}$ & $\textbf{92.2}$  & $\textbf{85.4}$ & $\textbf{90.3}$ & $\textbf{90.9}$ & $\textbf{84.2}$ & $\textbf{91.0}$ & $\textbf{71.5}$ & \textbf{83.2} \\
\toprule
\end{tabular}} 
\vspace{-5pt}
\label{pruning}
\end{table*}

\paragraph{Difference with Model Pruning.}
\label{diff}

Mode partition maintains important dynamic parameters while making redundant ones static, which may be similar to network pruning. In Table \ref{pruning}, we compare mode partition with network pruning \citep{lee2018snip} on the GLUE benchmark for BERT-base and reveal their difference empirically. PAD-Net achieves the best performance among all tasks listed, with 1.2\% average improvements over vanilla MoE. In contrast, we discover that network pruning lowers the performance of MoE significantly by 1.1\% on average. Considering maintaining the performance of a fully dynamic network, it is preferable to convert unimportant dynamic parameters into static ones than to prune them. 
\begin{figure}[t]
\caption{\textbf{Dynamic property calculation.} We plot layer-wise curves of parameter variance and output variance for ResNet-50. }
\centering
    \begin{subfigure}[t]{0.485\linewidth}
        \centering
        \includegraphics[width=\linewidth]{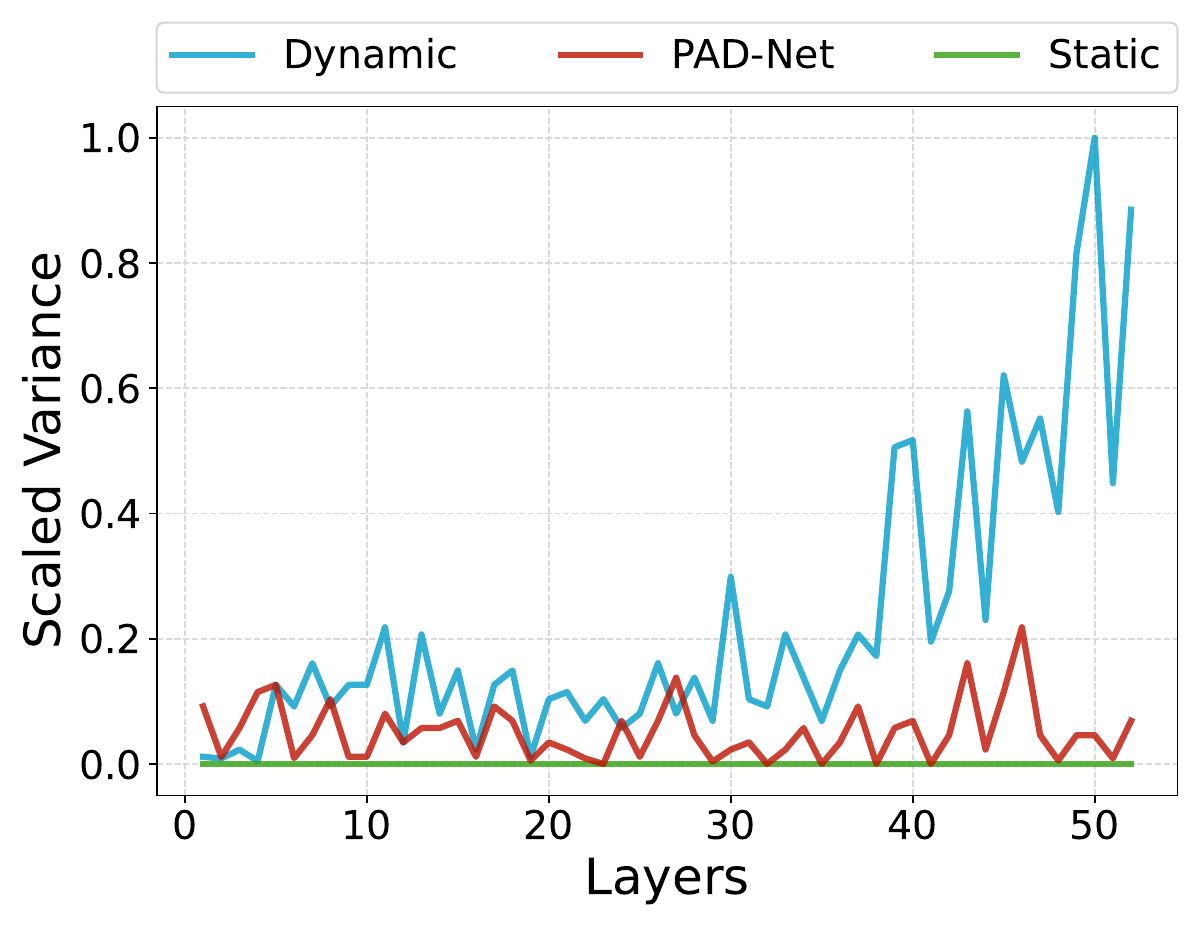}
        \subcaption{Parameter Variance}
    \end{subfigure}
    \begin{subfigure}[t]{0.485\linewidth}
        \centering
        \includegraphics[width=\linewidth]{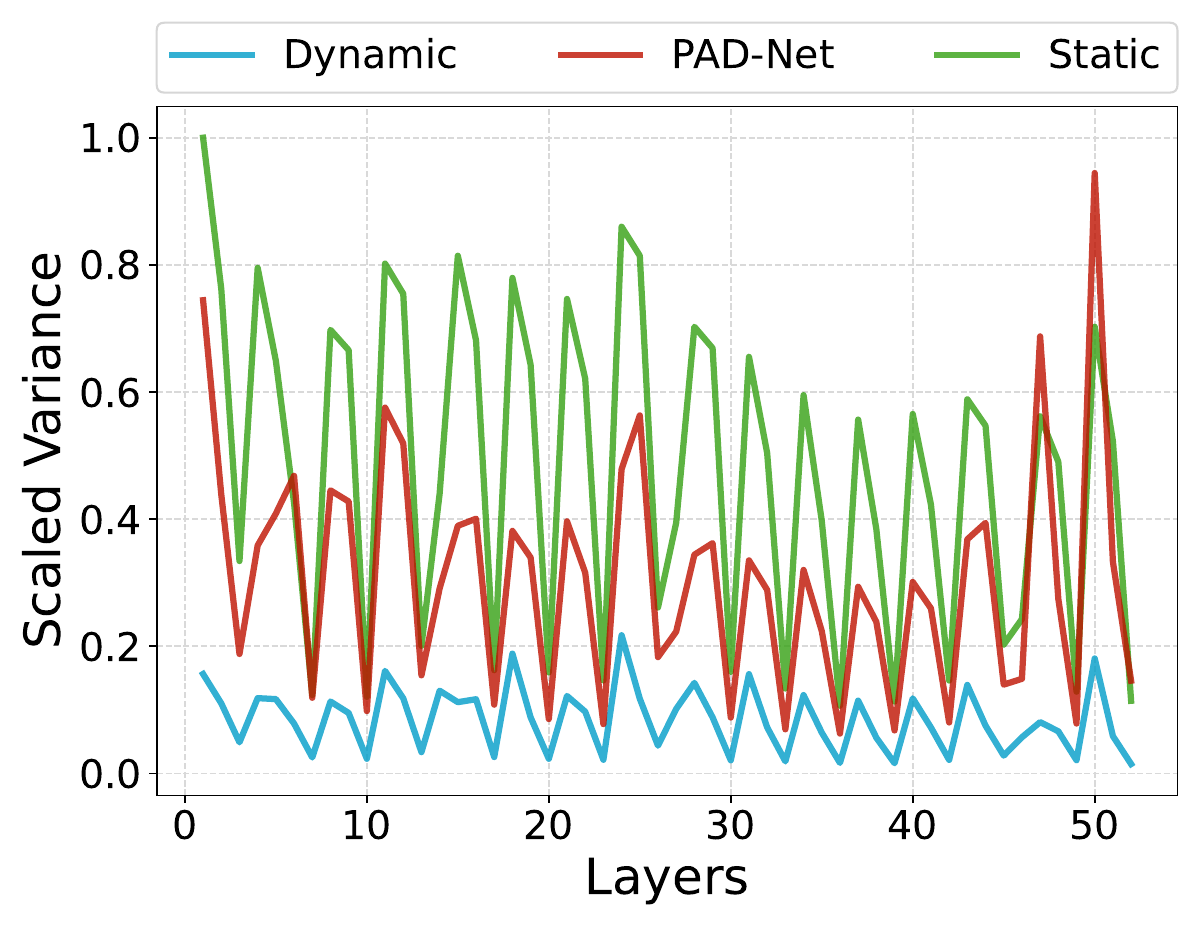}
        \subcaption{Output Variance}
    \end{subfigure}
\label{fig:dynamic_property}
\end{figure}
\paragraph{Dynamic Property. } Dynamic property refers to variant numerical characteristics of a dynamic network caused by different input samples. The ideal dynamic network maintains two capacities: assigning specific parameters for the input and making counterpart output discriminative. Inspired by \citet{li2021revisiting}, we take two levels of variance as metrics (parameter variance and output variance) to measure the dynamic property and show the result in Figure \ref{fig:dynamic_property}. Static convolution, dynamic convolution, and PAD-Net ($\kappa = 30\%$) show different properties given the same samples from ImageNet. We see that dynamic convolution retains a high degree of parameter variance while it has the lowest output variance. Static convolution performs the opposite. The outputs of PAD-Net are discriminative, which may contribute to its superiority in performance. 


\section{Conclusion and Future Work}
\label{sec:conclusion}
In this work, we first reveal parameter redundancy and high deployment costs of fully dynamic networks. To resolve these problems, we proposed the partially dynamic network (PAD-Net) to advance both performance and efficiency. PAD-Net demonstrated its superiority on MoE and DY-Conv frameworks. Extensive experiments on both NLP and CV tasks empirically show its effectiveness and efficiency against fully dynamic networks, significantly improving performance with much fewer dynamic parameters and less computation. Our proposed method could be extensively integrated with other mainstream architectures and inspire future work in efficient neural network designation and other fields.  
\section*{Acknowledgements}
We are grateful to the anonymous ACL reviewers and the area chair for their insightful comments and suggestions.

\section{Limitations}
\label{:sec:Limitations}
Despite the progress we made, there still exist limitations in our work. On the one hand, we only investigated some classic dynamic networks and found that the proposed method contribute to the best performance in selected criteria. However, other advanced partition methods that further improve the performance and efficiency may exist, which deserve exploration in future work. On the other hand, since we only consider MoE and DY-Conv in limited tasks, it would be valuable to consider other architectures (e.g., Switch Transformer~\cite{fedus2021switch}), machine learning methods (e.g., reinforcement learning~\cite{DBLP:journals/taslp/LiPLWNWYW22}) and tasks (e.g., machine translation~\cite{ding-etal-2020-self,ding2021improving}). 

\section*{Ethics Statement}
We take ethical considerations seriously and strictly adhere to the ACL Ethics Policy. This paper focuses on the higher efficiency of dynamic networks, e.g., the mixture of experts. Both the datasets and models used in this paper are publicly available and have been widely adopted by researchers. We ensure that the findings and conclusions of this paper are reported accurately and objectively.

\bibliography{custom_without_url}

\bibliographystyle{acl_natbib}

\end{document}